\pdfoutput=1
\documentclass[11pt]{article}

\usepackage{EMNLP2023}

\usepackage{graphicx}

\usepackage{times}
\usepackage{latexsym}
\usepackage{amsmath}
\usepackage{amssymb}
\usepackage{comment}
\usepackage{tabularx,ragged2e}
\usepackage{booktabs, multirow}
\usepackage{array, makecell, caption}

\newcolumntype{C}{>{\Centering\arraybackslash}X} % centered "X" column
\newcommand{\red}[1]{\textcolor{red}{#1}}
\newcommand{\blue}[1]{\textcolor{blue}{#1}}
\newcommand{\mat}[1]{\mathbf{#1}}

\usepackage[ruled, lined, linesnumbered, commentsnumbered, longend]{algorithm2e}

\usepackage[T1]{fontenc}
\usepackage[utf8]{inputenc}

\usepackage{microtype}

\usepackage{inconsolata}

\title{Unsupervised Learning of Graph from Recipes}

\author{A\"issatou Diallo\thanks{\quad Corresponding author: \texttt{a.diallo@ucl.ac.uk}\\}, \quad Antonis Bikakis, \quad Luke Dickens, \quad Rob Miller, \quad Anthony Hunter \\
University College London, United Kingdom \\}

\begin{document}
\maketitle

\begin{abstract}
Cooking recipes are one of the most readily available kinds of procedural text. They consist of natural language instructions that can be challenging to interpret.  In this paper, we propose a model to identify relevant information from recipes and generate a graph to represent the sequence of actions in the recipe.  In contrast with other approaches, we use an unsupervised approach. We iteratively learn the graph structure and the parameters of a $\mathsf{GNN}$ encoding the texts (text-to-graph) one sequence at a time while providing the supervision by decoding the graph into text (graph-to-text) and comparing the generated text to the input. We evaluate the approach by comparing the identified entities with annotated datasets,  comparing the difference between the input and output texts,  and comparing our generated graphs with those generated by state of the art methods.
\end{abstract}

\section{Introduction}
\label{sec:intro}

Procedural texts are a common type of natural language writing that provides instructions on following a procedure. They are essential in everyday life as they enable us to learn skills, complete tasks, and solve problems. Generally, they are loosely structured in step-by-step format with temporal order (as they would occur in real life) and logical flow (the results of previous steps feed into later steps). A procedural text can be used to guide an intelligent system for a wide range of applications, such as automated robotics, but the representation must be in a form that a computer-based system can interpret \cite{papadopoulos_learning_2022}. Given this, these representations must ideally capture all relevant entities and actions from the original text, while encoding all relationships that exist between them. A graph is a natural choice for this purpose as it represents not only the attributes of the entities but also their relationships.  
Moreover, being able to represent procedural text within a graph structure can facilitate reasoning and inference, such as adapting procedures when resources are unavailable or actions are impossible, see \cite{bikakis_graphical_2023}.

More precisely, procedural text understanding implies correctly identifying the resources and actions effects on resources, as well as understanding the flow between the procedures. For example, given the recipe \textit{"combine flour, sugar and milk in a bowl. Then, add the chocolate chips to the mixture. Finally, bake the cookies at 250 degrees."}, an intelligent system must be able to: (i) correctly identify entities and their types including implicit entities, e.g. "mixture" as a combination of previously mentioned ingredients; (ii) understand the temporal flow of the instructions, e.g. one must combine the first three ingredients into a batter before adding the chocolate chips; (iii) recognize the logical flow of information, e.g. the "bake" action acts on the output of the"combine" and "add" actions. This presents a challenge not present in factoid based reading comprehension \cite{clark_simple_2018}, in that our putative procedural text interpreter must model the dynamics and causal effects not explicitly mentioned in the text at hand \cite{dalvi_tracking_2018,bosselut_simulating_2018}.

\begin{figure}[!t]
    \centering
    \includegraphics[width=7.5cm]{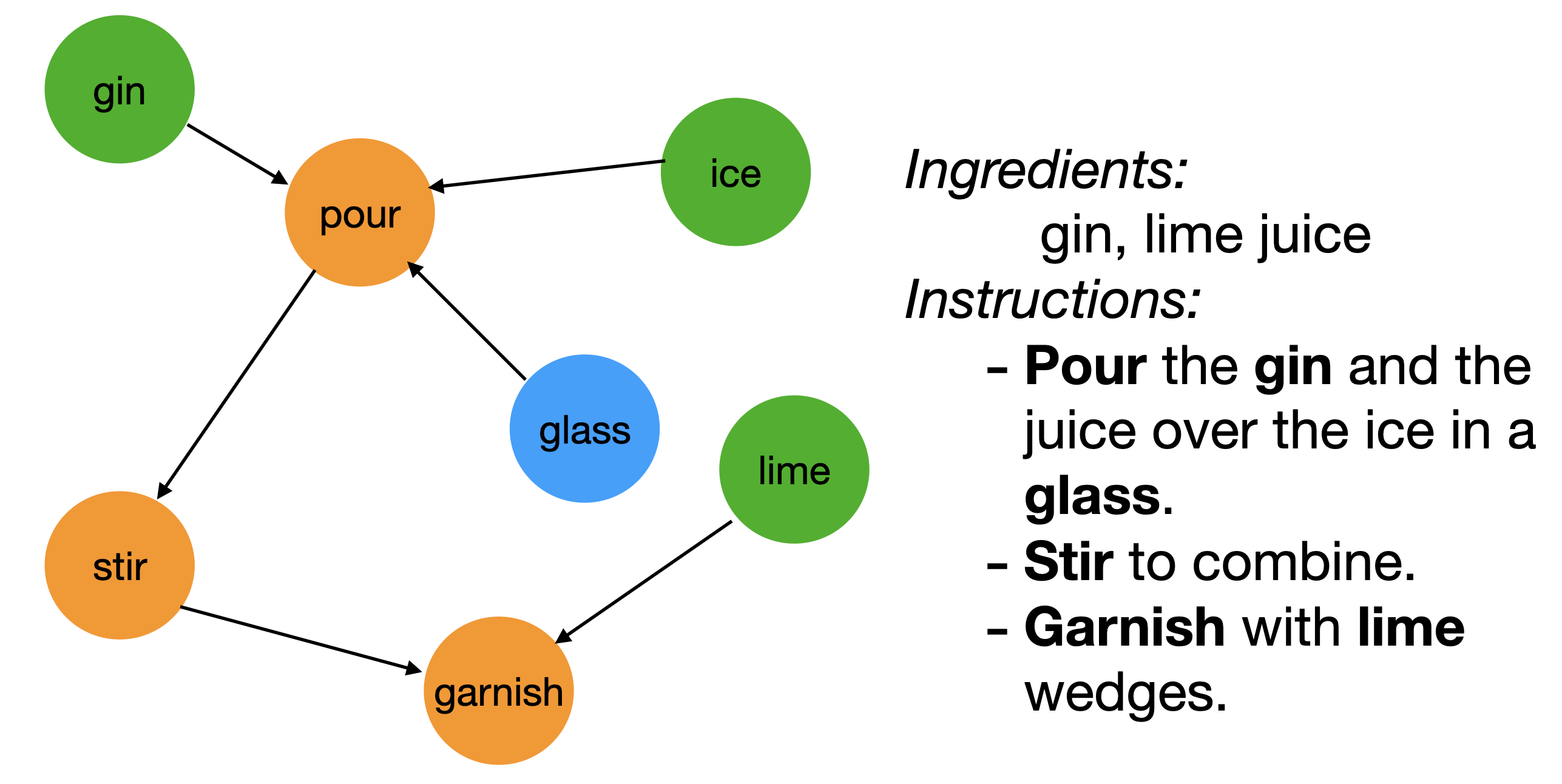}
    \caption{Example of output graph with the associated recipe. The identified entities are in bold.}
    \label{fig:entity_identifier}
    \vspace{-0.5cm}
\end{figure}

This paper presents a model, trained in a self-supervised fashion to encode cooking recipes as graphs, that identifies key actions, entities, including those that are implicit, and their relationships. We conduct a series of experiments showing that our proposed method yields meaningful graphs which can be used to reason about the original recipe. Our key contributions can be summarized as follows: (i) proposing a scalable approach to transform procedural texts into graphs while learning discrete and sparse dependencies among the entities using as a case study the cooking domain, and (ii) presenting a new method for jointly learning the edge connectivity of a graph in absence of explicit supervision at the graph-level.
%The paper is organized as follows

\section{Background}
In this section, we introduce some preliminary notions on graph theory and graph neural networks (GNNs). % and we formalize the problem of graph structure learning from procedural texts such as cooking recipes.

\paragraph{Graph Theory Basics}

A (heterogeneous) graph $\mathcal{G}$ can be described by the tuple $\mathcal{G}=\{\mathcal{V},\mathcal{E}\}$, where $\mathcal{V}=\{v_1, ..., v_N\}$ is a set of $N$ nodes and $\mathcal{E}=\{e_1, ..., e_M\}$ is a set of $M$ edges. In the case of heterogeneous graphs, there is a node type mapping function $\phi : \mathcal{V} \rightarrow \mathcal{C}$. 
For a given graph $\mathcal{G}$, the corresponding adjacency matrix $\mathbf{A} \in \{0,1\}^{N \times N}$ describes the connectivity between the nodes. The connected nodes are called neighbour nodes. $\mathbf{A}_{i,j}=1$ indicates that $v_i$ is adjacent to $v_j$, otherwise $\mathbf{A}_{i,j}=0$. Following the notation of \citet {franceschi_learning_2019}, we denote the set of all $N \times N$ adjacency matrices as $\mathcal{H}_N$. 

\paragraph{Graph Convolutional Networks}
GNNs are machine learning models designed for reasoning with graph-structure data while modelling relational and structural information by learning neighbour-aware node representations. In this study, we use Graph Convolutional Network ($\mathsf{GCN}$) \cite{kipf_semi-supervised_2017} to aggregate the neighbour information for each entity node.
%\vspace{-0.25cm}
%and its  
%standard graph learning function is $\mathbf{H}^{l+1} = f(\mathbf{H}^{l}, \mathbf{A})$,
%\begin{equation}
%    \mathbf{H}^{l+1} = f(\mathbf{H}^{l}, \mathbf{A})
%\end{equation}
%where $\mathbf{H}^l$ and $\mathbf{H}^{l+1}$ are the updated node embeddings at the $l^{th}$ layer of the network. $f$ indicates the message aggregation function. In GNN, each node $i$ aggregates the critical information of its neighbours $N(i)$ at a given layer $l$ through $f$. 
\begin{equation}
    \mat{h}^{l+1}_i= \sigma \left( \mat{W}\sum_{j \in N(i)\cup \{i\}} \frac{1}{\sqrt{d(i)d(j)}}\mat{h}^l_i \right)
\label{eq:gcn_eq}
\end{equation}
Eq. \ref{eq:gcn_eq} describes the propagation mechanism for the $\mathsf{GCN}$. At layer $l$, each node $i$ aggregates the representation of itself and all its neighbouring nodes $N(i)$ based on $\mat{A}$. Then, the updated representation $\mat{h}^{l+1}_i$ is computed based on the weight matrix of the layer $\mat{W}$, the activation function $\sigma$, normalized by the degree of the source $d(i)$ and its connected nodes $d(j)$.

\section{Model}

\begin{figure*}[ht]
    \centering
    \includegraphics[width=16cm]{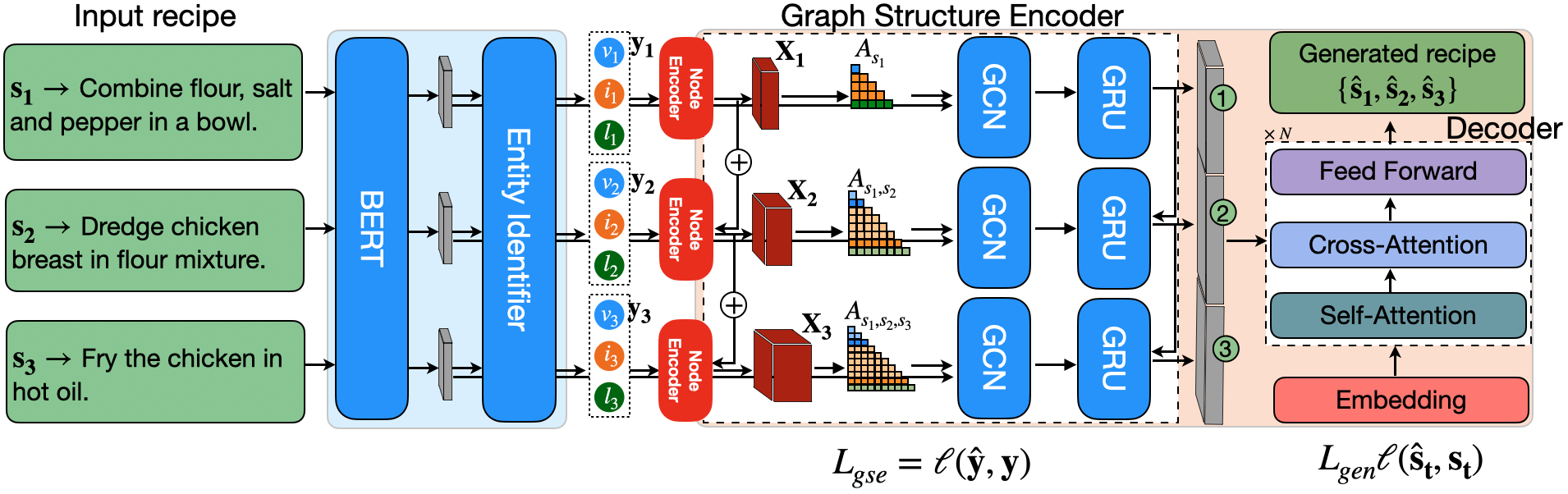}
    \caption{Overview of the proposed model. The first part, in blue, encodes the procedures and identifies the relevant entities (refer to Figure \ref{fig:entity_identifier} for more details). These are passed to the second part (in red) containing the Graph Structure Encoder module, which learns the adjacency matrix and the graph encoding (with cost $L_{gse}$) (refer to Figure \ref{fig:recgraph}) and the decoder that back-translate the graph into a recipe (with cost $L_{gen}$). }
      
    \label{fig:recipe2graph}
\vspace{-0.25cm}
\end{figure*}

In this paper, we address the challenge of producing graphs of cooking recipes when no supervision in terms of graphs is available. 
%For this purpose, we aim to learn a discrete and sparse graph structure between three different types of nodes: ingredients, cooking actions and locations.
We aim to discover implicit entities and relationships between actions, ingredients and locations, and for the relationships to carry a similar meaning, e.g. an ingredient relates to an action that acts upon it without explicitly enforcing them so that the output is loosely aligned with the semantics of flow-graphs such as \cite{mori_flow_2014}.
%For doing this, three sets of predefined entities that need to be tracked are given a priori, $\mathcal{V}=\{v_1, ..., v_V\}$. 

%\section{Method}

%In this paper, we address the challenge of producing graphs using cooking recipes when no graph level supervision is available.
%In contrast to \cite{mori_flow_2014}, 

\paragraph{Task Definition} 
A recipe $R$ consists of $T$ sentences $R=\{s_0, s_1, ..., s_T\}$. Each sentence $s_i$ contains the action $a_i$ that operates on $i_i$,
 a set of ingredients in (a set of) $l_i$ locations. All cooking actions, ingredients and locations are part of a pre-specified set of entities. Often, a cooking action can lead to an intermediate output such as \textit{batter, dough} or \textit{mixture}. In the scope of this work, these intermediate outputs are considered ingredients. The model aims to learn a graph describing the interactions between actions, ingredients and locations for a given recipe. These entities are represented as nodes in the output graph and the edges are the connections between the $|\mathcal{C}|=3$ different types of nodes, namely \textit{actions, ingredients} and \textit{locations}.
 %through the node type mapping function $\phi : \mathcal{V} \rightarrow \mathcal{C}$. 
 In the absence of a dataset containing pairs of recipes and their corresponding cooking graphs, we tackle this problem in an unsupervised way.

\paragraph{Overview} For doing this, given a recipe we first parse it such that each instruction contains only one cooking action (e.g., "\textit{sprinkle with salt // mix well // set aside in a bowl}"). Then, a previously trained model is applied to each instruction to encode the sentence and to identify the relevant entities that should be present in the cooking graph. Next, these elements are sequentially passed to a graph encoder that jointly learns the graph structure as an adjacency matrix, the node and graph representations. The resulting graph encoding is used as a conditioning for a transformer-based module that translates it back to textual cooking instructions. Finally, the output text is compared to the original set of cooking instructions in a cyclic way to guide the learning process. 

\subsection{Entity Identifier} Given a sentence $s_t$ with $L_s$ tokens such that $s_t=\{w_{i,1}, ... , w_{i,L_s\}}$, we use $\mathsf{BERT}$ \cite{devlin_bert_2019} as a context encoder, which is based on a stack of multilayer \textit{transformer blocks} \cite{vaswani_attention_2017}. 
The instruction representation $\mat{h}_{s_t}$ is obtained by extracting the hidden state $\mat{h}^{s_t}_0$ corresponding to the special token $[\texttt{CLS}]$ so that $\mat{h}_{s_t} = \mat{h}^{s_t}_{[\texttt{CLS}]}$:
\begin{equation}
    %\mathbf{h}_{s_t}=BERT\{s_t\}
    \{\mat{h}^{s_t}_{[\texttt{CLS}]}, ..., \mat{h}^{s_t}_{L_s}\} = \texttt{BERT}\{[\texttt{CLS}], ...,w^{s_t}_{L_s} \}
\label{eq:bert_representation}
\end{equation}

\label{sec:entity_id}
The first block of our model 
%has the purpose of interpreting instructions written in natural language as sequences of processes of actions and their overall effects on ingredients and locations as entities. Specifically, this block, 
takes as input one sentence at a time and extracts which actions are executed on the ingredients while tracking the changes in location and the transformation of the ingredients. 
Existing procedural text understanding models rely on some type of name entity recognition sub-task to obtain the entities the model should focus on such as in \cite{tang_procedural_2022}. This strategy would be sufficient if entities were always explicitly
mentioned, but natural language often elides these arguments or uses pronouns. As such, the module
must be able to consider entities mentioned previously or under different names. 

\begin{figure}[ht]
    \centering
    \includegraphics[width=7.49cm]{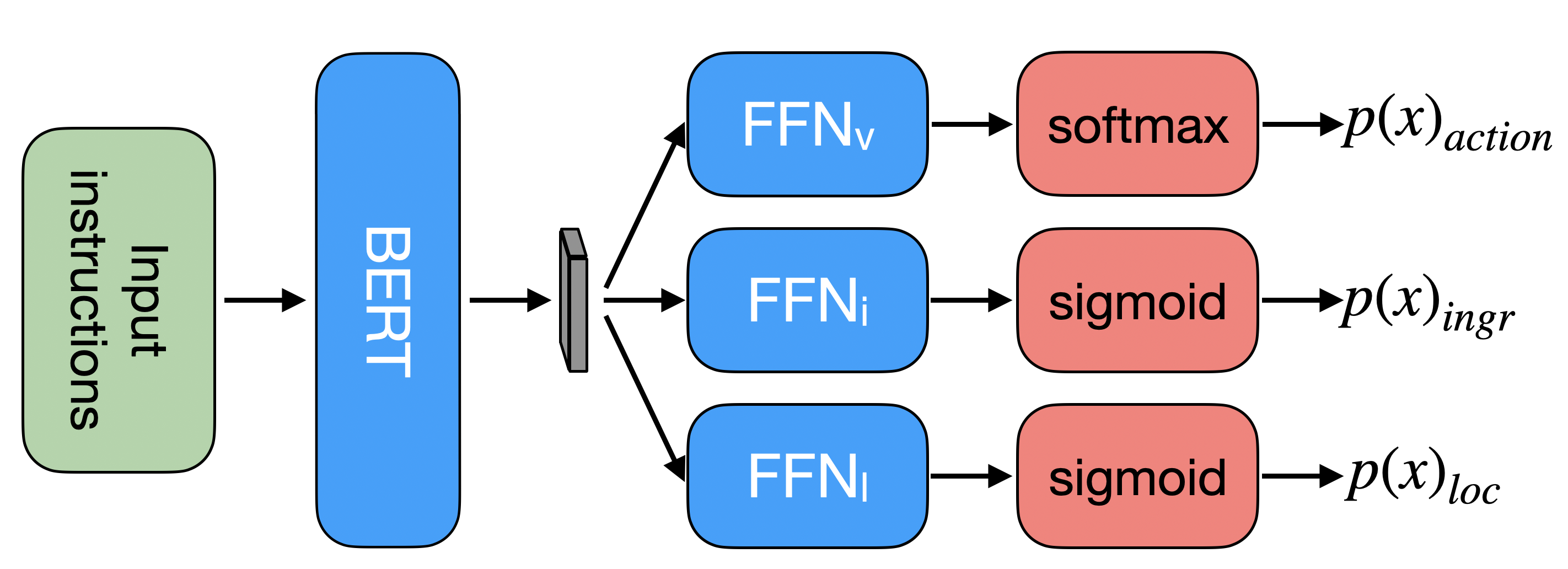}
    \caption{Details of the Entity Identifier. The recipe is parsed such that each sentence contains only one action whereas multiple ingredients and locations are permitted.}
    \label{fig:entity_identifier}
\vspace{-0.25cm}
\end{figure}

This problem can be seen as a multitask learning problem where we need to learn to identify actions, ingredients and locations simultaneously. 
Specifically, a feed-forward network (FFN) with a cross entropy loss is used to identify the correct cooking action. Concurrently, two FFNs with a binary cross entropy cost function are used to learn the subsets of ingredients and locations for a given instruction $s_t$. Figure \ref{fig:entity_identifier} illustrates this component of our approach. 
%We describe here the process for the ingredients but a similar procedure is applied to obtain the actions and locations. Let us define the pre-specified list of ingredients of size $F$ as $L_F=\{f_i\}^F_{i=0}$. Each sentence contains the mention of $k_{ingr}$ ingredients entities from $L$ forming a subset $C_{ingr}$. We encode this subset of size $k_{ingr}$ as a binary vector using one-hot encoding. In this scenario, the goal is to predict $\hat{C}_{ingr}^{(i)}$ from an instruction $s^{(i)}$ 
%by maximising the following objective 
%\begin{equation}
%    \argmax_{\theta}\sum^T_{i=0}\log p (\hat{C}_{\cdot}^{(i)} = C^{(i)}|s^{(i)};\theta)
%\end{equation}
%where $\theta$ represents the learnable parameters of a simple feed-forward network with a sigmoid activation.

\subsection{Graph Structure Encoder}
\label{sec:gse}
We aim to learn the adjacency matrix and the graph representation of a graph $G$ associated with a recipe $R$ and a set of (heterogeneous) entities. Our encoder takes as input one instruction at a time until the end of the recipe. It recursively generates the graph for the block of the cooking recipe seen up to the $t$-th generation step. The output is conditioned on the already generated graph representation at time step $t-1$. 
The process is divided into two main parts: (1) generating the structure of the graph, described by the adjacency matrix; (2) learning the parameters for the $\mathsf{GCN}$ as well as obtaining the final graph representation. There are two main approaches for generating the adjacency matrix. The first is to think of the problem as a probabilistic one and to learn the parameters of a probability distribution over the graphs as in \cite{franceschi_learning_2019}. The second method is to work with a continuous relaxation of the adjacency matrix. We chose the latter approach in this work. 

 We recall that the node features of the entities are not given and need to be learned along the graph structure. For doing this, we initialize the feature matrix $\mathbf{X}$ containing the node representations with a pre-trained model \cite{pellegrini_exploiting_2021} specific to the cooking domain, acting as node encoder. More specifically, given a recipe with $T$ instructions each with an action $a_i$, $i_i$ ingredients and $l_i$ locations with $i \in \{0,...,T\}$, we first concatenate the indices denoting the entities into a single list of indices. We impose a specific ordering for the concatenation: [\textit{action; ingredients; locations}] and within each subset, the items are sorted in alphabetical order. This yield 
 %the feature matrix 
 $\mathbf{X}^{(t)}$ for the $t$-th sentence in the recipe with the initialized embedding layer described earlier. We now explain the details of the graph generation step below. 

 \paragraph{Relation matrix} We extract a relation matrix $\mathbf{R}^{(t)}$ characterizing the probability of a relation (an edge) between two nodes based on $\mathbf{X}^{(t)}$. This is meant to capture the relationship between the nodes and propagate this relation to the topological structure of the graph. We first apply two different mappings to $\mathbf{X}^{(t)}$ to obtain a $\mathbf{Z}^{(t)} = \textsf{FFN}(\mat{X}^{(t)})$ and $\mathbf{M}^{(t)} = \textsf{FFN}(\mat{X}^{(t)})$. 
 %\begin{align}
 %    \mathbf{Z}^{(t)} = W_1\mathbf{X}^{(t)} + b_1 \; \text{and} \;
 %    \mathbf{M}^{(t)} = W_2\mathbf{X}^{(t)} + b_2 
 %\end{align}
Finally, the relation matrix is obtained as $\mathbf{R}^{(t)} = \mathbf{Z}^{(t)}\mathbf{M}^{(t)^\top}$. 

\paragraph{Adjacency Processor} The relation matrix $\mathbf{R}^{(t)}$ is not suited to be used as an adjacency matrix, as it may contain both positive and negative values and it is non-normalized. Let us define a function $\mathcal{S}$ that takes as input the relation matrix $\mathbf{R}^{(t)}$ and outputs a continuous relaxation of the adjacency matrix $\mathbf{A}^{(t)}$. $\mathcal{S}$ will take a matrix and transform it into a low entropy one, meaning that most of its elements will be equal to zero, hence augmenting its sparsity. A simple algorithm for this purpose is the Sinkhorn-Knopp algorithm \cite{knopp_concerning_1967}, which consists in iteratively rescaling the rows and columns of a square matrix with positive entries. By applying the algorithm in the log domain, we can apply the algorithm to the relation matrix and obtain an adjacency matrix by finally applying the $\exp$ operator to the output for getting positive values. To sum up, $\mat{A}^{(t)} = \mathcal{S}(\mat{R}^{(t)})$\footnote{For the sake of brevity, we slightly abuse the notation because the $\mathcal{S}(\mat{R}^{(t)})$ is not a  $(0,1)$-matrix but a continuous relaxation.}.

\paragraph{Node Classification}  
%After obtaining the adjacency matrix $\mat{A}^{(t)}$ at the $t$-th step, we need to normalize it before using it with the GNN to make predictions about the graph structure. The normalized adjacency matrix $\mathbf{\tilde{A}}$ is defined as: $\mathbf{\tilde{A}} = \mathbf{D}^{-\frac{1}{2}}\mathbf{A}\mathbf{D}^{-\frac{1}{2}}$.
 
We choose a two-layer $\mathsf{GCN}$ \cite{kipf_semi-supervised_2017} that takes as input the feature matrix $\mathbf{X}^{(t)}$ and the adjacency matrix $\mat{A}^{(t)}$. The first layer of the GCN architecture projects the inputs into an intermediate latent space and the second layer maps it to the output embedding space.
Due to the absence of labelled graphs to guide the learning process, we include a \textit{pretext task}, which is a pre-designed task for the network to solve such that it can be trained by the learning objective functions
of the pretext task and the features and parameters are learned through this
process. 
A natural choice for the pretext task is node classification
%, hence the output embedding space has dimensionality $|\mathcal{C}|$ 
%as follows $\mathsf{GCN}: \mathbb{R}^{d} \times \mathbb{R}^{h} \rightarrow \mathbb{R}^{|\mathcal{C}|}$. 
and the goal is to predict the entity type of each node by providing the logits for each class through $\mathrm{softmax}$ in this way $\mathbf{\hat{y}} = \mathrm{softmax}(\mathsf{GCN}(\mat{X}, \mat{A}))$. 
\begin{gather}
%\mathbf{H} = \mathrm{ReLU}(\mathbf{\tilde{A}}\mathbf{X}W_3) \\
%\mathbf{\hat{y}} = \mathrm{softmax}(\mathbf{\tilde{A}}\mathbf{H}W_4) \\
\mathcal{L}_{gse} = \ell(\mathbf{\hat{y}}, \mathbf{y}) \label{eq:encoder_loss}
\end{gather}
where $\ell$ corresponds to a cross-entropy loss function, $\mathbf{y}$ are the ground-truth labels known before-hand and $\mathbf{\hat{y}}$ are the predicted labels. Eq. \ref{eq:encoder_loss} will constitute the first term of the global loss.   

\begin{algorithm}[t]
    %\SetKwFunction{isOddNumber}{isOddNumber}
    % \SetKwInput{Input}{Input}
    % \SetKwInput{Output}{Output}
    \SetKwInOut{KwIn}{Input}
    \SetKwInOut{KwOut}{Output}

    \KwIn{ Trained model, \\ $R=\{s_0, ..., s_T\}$: the input recipe;}
    \KwOut{ $\mat{A}$: adjacency matrix; \\ $Y$: nodes; \\ $Y_{\mathcal{C}}$: nodes classes; }
    $t \leftarrow 0$ \\
     $graphSequence = [\ ]$
    
    \For{$s_t$ in $R$}{
        $\mathbf{X}^{(t)}$, ${Y}^{(t)} \leftarrow $ Entity Identifier$(s_t)$ \\
        $\mathbf{Z}^{(t)}$, $\mathbf{M}^{(t)} \leftarrow $ $W_1\mathbf{X}^{(t)} + b_1$, $W_2\mathbf{X}^{(t)} + b_2$ \\
        $\mathbf{R}^{(t)} \leftarrow \mathbf{Z}^{(t)}\mathbf{M}^{(t)}\top$ \\
        $\mathbf{A}^{(t)} \leftarrow $ Adjacency Processor$(\mathbf{R}^{(t)})$ \\
        $\tilde{\mathbf{A}}^{(t)} \leftarrow $ Normalization of $\mathbf{A}^{(t)}$ \\
        $\tilde{\mathbf{G}}^{(t)} \leftarrow \mathsf{GCN}(\mathbf{X}^{(t)}, \tilde{\mathbf{A}}^{(t)})$ \\
        $\tilde{\mathbf{g}}^{(t)} \leftarrow $ Graph pooling $(\tilde{\mathbf{G}}^{(t)})$ \\
        ${graphSequence}.append(\tilde{\mathbf{g}}^{(t)})$ \\       
    }
    $\mathbf{G} \leftarrow \mathsf{GRU}(graphSequence)$

    \KwRet{$\mathbf{G}$, $\mathbf{A}^{(T)}$, ${Y}^{(T)}$, ${Y}^{(T)}_{\mathcal{C}}$}
    \caption{Graph Structure Encoder}
    \label{alg:algo}
\end{algorithm}

\paragraph{Recurrent Graph Embedding}
We apply the steps detailed above iteratively until reaching the end of the recipe. At each time step, the node vectors are the concatenation of the unique node indices seen up to the $t$-th sentence of the recipe such that the last iteration, the adjacency matrix $\mathbf{A}^{(t=T)}$ has $m$ unique nodes with $m=a_T + i_T + l_T$. Each graph encoding $\mathbf{H}^{(t)}$ goes through a mean pooling function in order to obtain a global representation as a single graph embedding. 
These vectors are then stacked into a matrix $\mathbf{\tilde{G}}^{R} \in \mathbb{R}^{T \times d}$. We concatenate $\mathbf{\tilde{G}}^{R}$ to the sentence representations from eq. \ref{eq:bert_representation} to obtain $\mathbf{{G}}^{R} \in \mathbb{R}^{T \times 2d}$.

\begin{figure}[ht]
    \centering
    \includegraphics[width=7.25cm]{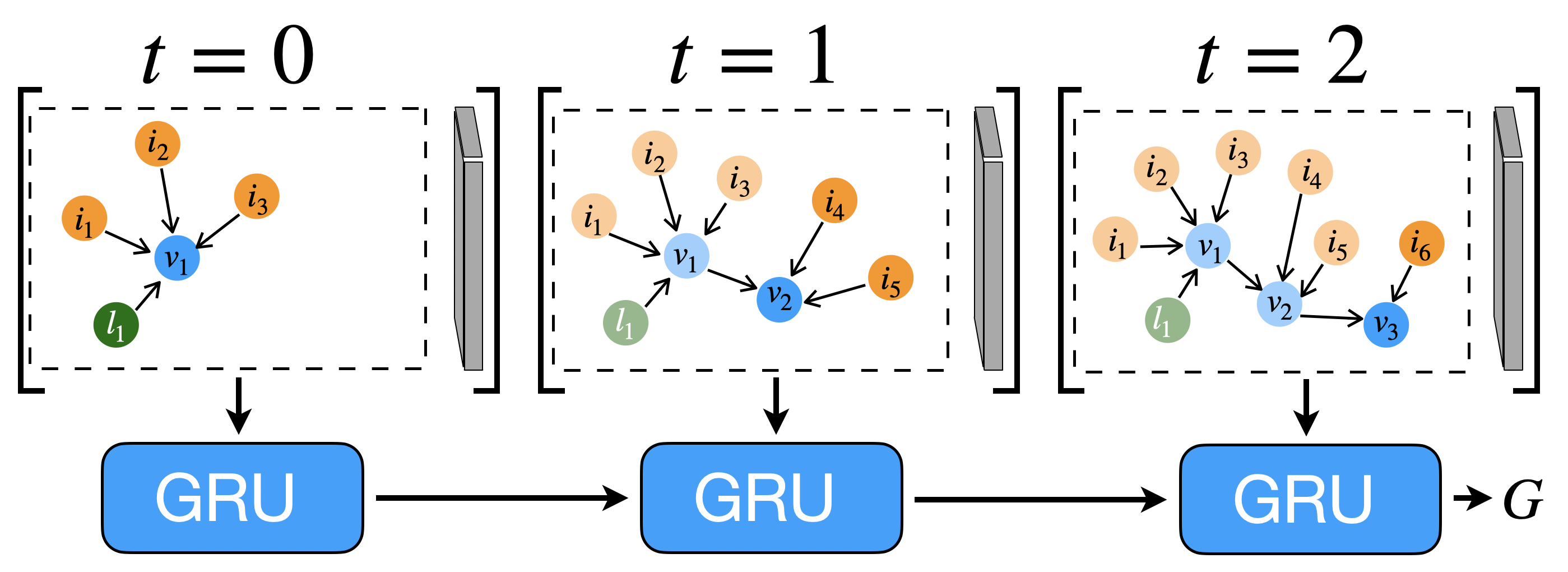}
    \caption{We exploit the ability of $\mathsf{GRU}$ at handling temporal dependencies to  increase the expressive power of the iteratively constructed partial graphs. The $\mathsf{GRU}$ has two inputs: the current input (the (partial) graph at $t$ concatenated with $\mat{h}_{s_t}$) and the previous state (output of a $\mathsf{GRU}$ at $t-1$). %The previous state is the output of a $\mathsf{GRU}$ at time $t-1$ whereas the current input is the current (partial) graph at time step $t$. 
    }
    \label{fig:recgraph}
\vspace{-0.25cm}
\end{figure}

The final sequence of graph representations is obtained using a two-layer bidirectional $\mathsf{GRU}$ to update the input sequence as seen in Figure \ref{fig:recgraph}. Algorithm \ref{alg:algo} describes the graph generation process. Finally, this sequence is used to condition the transformer-based decoder. 

\subsection{Cooking Instruction Decoder}
\label{sec:decoder}
Given a graph sequence representing a recipe with $T$ steps, the goal of the cooking instruction decoder is to produce a sequence of textual instructions $\hat{R}=\{\hat{s}_0, \hat{s}_1, ..., \hat{s}_T\}$ by means of a transformer-based architecture. This decoder is conditioned on the sequence of graph embeddings provided by the graph encoder described earlier. 

The instruction decoder is composed of \textit{transformer blocks} each of them containing a self-attention layer applying self-attention over previously generated outputs, a cross-attention layer that attends to the model conditioning in order to refine the self-attention output and a linear layer. This process is the
same as the original Transformer \cite{vaswani_attention_2017} and we omit it due to the limited space. A final $\mathrm{softmax}$ provides a distribution over the words for each time step. We use the cross-entropy loss $\mathcal{L}_{gen}$ between the generated instructions $\hat{s}_t$ and the original instructions ${s}_t$. 

\subsection{Optimization}
We train our model, outlined in Figure \ref{fig:recipe2graph} in two stages.  In the first phase, we train the entity identifier described in section \ref{sec:entity_id}. Then, in the second stage, we train the graph structure encoder from section~\ref{sec:gse} and the cooking instruction decoder from section~\ref{sec:decoder}. The total loss used to perform the learning is $\mathcal{L}_{tot}= \mathcal{L}_{gse} + \mathcal{L}_{gen} + \lambda\|A\|_1$. The last term is a regularization term that encourages the sparsity of the adjacency matrix. We apply a bi-level optimization approach by alternatively updating the weights involved in the learning of the adjacency matrix and the weights for the $\mathsf{GNN}$. The cooking instruction decoder is trained with teacher forcing \cite{williams_learning_1989}.

\begin{comment}
\begin{table}[ht]
\begin{tabular}{@{}lll@{}}
\toprule
Model                                  & F1             & Recall         \\ \midrule
2-layer LSTM Entity Recognizer         & 50.98          & 13.33          \\
Adapted Gated Recurrent Entity  & 45.94          & 7.74           \\
EntNet       & 48.57          & 9.87           \\
Neural Process Networks                & 55.39          & 20.45          \\ \midrule
\textbf{Ours}                          & \textbf{65.14} & \textbf{63.44} \\ \bottomrule
\end{tabular}
\caption{Results for entity selection.}
\label{tab:entity_classification results}
\end{table}
\end{comment}

\section{Experiments}
\begin{comment}
\begin{table*}[]
\centering
\begin{tabular}{@{}lllll@{}}
\toprule
 & \multicolumn{2}{c}{\textbf{Now You're Cooking}} & \multicolumn{2}{c}{\textbf{English Flow Corpus}} \\ \cmidrule(l){2-5} 
 & \multicolumn{1}{c}{BLEU} & \multicolumn{1}{c}{ROUGE$_L$} & \multicolumn{1}{c}{BLEU} & \multicolumn{1}{c}{ROUGE$_L$} \\ \midrule
 NPN & 0.37 & 0.35 & - & - \\
Ours w/o s. & 0.42 & 0.41 & 0.44 & 0.39 \\
Ours & 0.51 & 0.45 & 0.49 & 0.38 \\ \bottomrule
\end{tabular}
\caption{Results for text $\rightarrow$ graph and graph $\rightarrow$ text.  Best model is identified by highest scores.}
\label{tab:gen_results}
\end{table*}
\end{comment}

\begin{table*}[!htb]
\begin{minipage}{.49\linewidth}
    \centering
    \medskip
\scalebox{0.85}{
\begin{tabular}{@{}lll@{}}
\toprule
Model                                  & F1             & Recall         \\ \midrule
2-layer LSTM Entity Recognizer         & 50.98          & 13.33          \\
Adapted Gated Recurrent Entity  & 45.94          & 7.74           \\
EntNet       & 48.57          & 9.87           \\
Neural Process Networks                & 55.39          & 20.45          \\ \midrule
\textbf{Ours}                          & \textbf{65.14} & \textbf{63.44} \\ \bottomrule
\end{tabular} }
\caption{Results for entity selection. The reported scores are averaged for all three types of entities.}
\label{tab:entity_classification results}

\end{minipage}\hfill
\begin{minipage}{.49\linewidth}
    \centering

    \medskip
\scalebox{0.85}{
\begin{tabular}{@{}lllll@{}}
\toprule
 & \multicolumn{2}{c}{\textbf{NYC}} & \multicolumn{2}{c}{\textbf{EFC}} \\ \cmidrule(l){2-5} 
 & \multicolumn{1}{c}{BLEU} & \multicolumn{1}{c}{ROUGE$_L$} & \multicolumn{1}{c}{BLEU} & \multicolumn{1}{c}{ROUGE$_L$} \\ \midrule
 NPN & 0.37 & 0.35 & - & - \\
Ours w/o s. & 0.42 & 0.41 & 0.44 & 0.39 \\
Ours & 0.48 & 0.45 & 0.49 & 0.38 \\ \bottomrule
\end{tabular} }
\caption{Text generation results for text $\rightarrow$ graph and graph $\rightarrow$ text.}
    \label{tab:gen_results}
\end{minipage}
\vspace{-0.25cm}
\end{table*}

For learning and evaluation, we use the "Now You're Cooking" (NYC) dataset used in \cite{bosselut_simulating_2018} that contains 65816 recipes for training, 175 recipes for validation and  700 recipes for testing. This dataset contains annotations for the ingredients, actions and locations for each sentence in a given recipe. For evaluating the quality of the graphs produced, we rely on the dataset "English Flow Corpus" (EFC) by \citet{yamakata_english_2020}.

\paragraph{Implementation details} We use the following settings. We train the entity identifier for 5 epochs with a learning rate of $2 \cdot 10^{-5}$. The nodes' hidden dimension is $100$. The 2-layer $\mathsf{GCN}$ outputs a graph representation of hidden dimension $768$. The graph structure encoder is trained with a learning rate of $10^{-3}$ and the instruction decoder has a base learning rate of $10^{-4}$ and a step decay of $0.1$ every 5 epochs. We train for 10 epochs. All experiments are conducted on a Nvidia Titan X GPU.

\subsection{Entity Selection} In this section, we evaluate the model performance in identifying the correct entities and modelling their interactions. This is framed as a classification task.  Following \citet{bosselut_simulating_2018}, we rely on the macro F1 score and the recall to measure the performance of the model. The Table~\ref{tab:entity_classification results} shows the results. A selected entity is denoted as one whose sigmoid activation is greater than 0.5. The reported scores are averaged for all three types of entities. 
 
\paragraph{Baselines} The main baseline for this stage of the framework consists of the Neural Process Networks by \citet{bosselut_simulating_2018}. For the sake of completeness, we report the scores of their baseline to better position our model. The first two baselines, represented by the first two lines of Table~\ref{tab:entity_classification results} consist of an $\mathsf{LSTM}$ and $\mathsf{GRU}$ model trained to predict the entities without any other external information. This is closer to our model. \textit{Adapted Recurrent Entity Network} \cite{henaff_tracking_2017} and \cite{bosselut_simulating_2018} are both $\mathsf{RNN}$-based models. The first one is a network doted with
external “memory chains” with a delayed memory update mechanism to track entities. The second one aims to simulate the
effects of actions on entities in procedural text. It embeds input sentences in a $\mathsf{GRU}$, then uses a $\mathsf{FFN}$ to predict actions which occur in each sentence and sentence-level
and recurrent attention mechanisms to identify entities affected by actions. 

\begin{table}[hb]
\centering
\begin{tabular}{@{}lc@{}}
\toprule
 & {Graph-edit distance} \\ \midrule
RandomGraph & 112.1 \\
Ours w/o sent & 71.4 \\
Ours & \textbf{67.1} \\ \bottomrule
\end{tabular}
\caption{Graph edit distance. Due to the high computational cost for graph matching, we report the result for 10 sample pairs of generated graphs and ground truths.}
\label{tab:graph_edit_distance}
\vspace{-0.45cm}
\end{table}

\paragraph{Results} From the results, we can observe that a transformer-based classifier outperforms all the baselines. However, it should be noted that in the ingredient type entity selection, the model sometimes fails. Given the sentence \textit{add sugar and remaining broth}, the model selects \textit{chicken broth and sugar} as ingredients while ignoring most of the other ingredients from the ground truth-set. This is probably due to the fact that there is a need to inject external information or explicitly track the ingredients that have been used and consumed throughout the recipe because the sole information contained in the sentence is not sufficient to infer a mapping to previous ingredients. The situation slightly improves, when including \textit{intermediates} "ingredients" as an entity that should be tracked. The model is not required to link the construct \textit{dough} to its single components but can continue to reason about actions and locations related to this food item. 

\begin{table*}[t]
\scalebox{0.9}{
\begin{tabular}{@{}ll@{}}
\textbf{Reference text} & \begin{tabular}[c]{@{}l@{}}add ingredients in the order suggested by your manufacturer. \\ set for dough setting and start the machine. when the unit signals remove dough.\\  pat dough into a greased 30cm ( 12 in ) round pizza pan. let stand 10 minutes. \\ preheat oven to 200 c / gas mark 6. spread pizza sauce over dough. \\ sprinkle toppings over sauce ( such as basil and mozzarella for a simple margherita pizza ). \\ bake 15 to 20 minutes, or until crust is golden brown.\end{tabular} \\
\midrule
\textbf{Generated text} & \begin{tabular}[c]{@{}l@{}}add the ingredients in the order listed. \\ set machine on, the dough cycle is best. \\ form the dough and pat into 12 - inch round pizza pan. \\ let stand 15 mins. preheat oven to 350f 200c. spread pizza sauce over dough. \\ sprinkle cheeses, then tomato sauce. \\ bake 10 to 15 minutes or until crust is golden brown\end{tabular} 
\end{tabular} }
\caption{Example of generated text from the graph representation of a recipe for making pizza.}
\label{tab:example_generated_text}
\end{table*}

\begin{table*}[ht]
\centering
\scalebox{0.89}{
\begin{tabular}{p{0.64cm}|p{5.95cm}|m{2.45cm}|m{5.95cm}} 
 \multicolumn{1}{c}{}& \multicolumn{1}{c}{} & \multicolumn{1}{c}{\textbf{Predicted}} & \multicolumn{1}{c}{\textbf{Ground truth}}  \\ \cline{1-4}
$s_{t-1}$ & if you use canned chickpeas, reduce or omit the salt. & \multirowcell{2}[0pt][l]{\blue{garlic}, \blue{yogurt}, \\ \red{chili}} & \multirowcell{2}[0pt][l]{\red{chili powder, curry powder}, \blue{garlic}, \\ \red{ginger, parsley, salt} , \blue{yogurt}} \\ \cline{1-2}
$s_t$ & mix yogurt , garlic and spices. &  &  \\ \cline{1-4}
$s_{t-1}$ & heat the oil. & \multirow{2}{*}{\red{beef}} & \multirowcell{2}[0pt][l]{\red{cornstarch, garlic clove, oil, oyster} \\ \red{sauce, pepper, sherry, soy sauce},\\ \red{sugar}} \\ \cline{1-2}
$s_t$ & stir-fry beef quickly, until the meat is medium rare. &  &  \\ \cline{1-4}
$s_{t-1}$ & dip bottom of balloon in melted chocolate and sit on cookie sheet. & \multirowcell{2}[0pt][l]{\blue{chocolate}} & \multirowcell{2}[0pt][l]{\blue{chocolate}, \red{raspberry}} \\ \cline{1-2}
$s_t$ & refrigerate until hard. &  &  \\ \cline{1-4}
$s_{t-1}$ & sift flour in a bowl. & \multirow{2.4}{*}{\blue{cornstarch, sugar}} & \multirow{2.4}{*}{\blue{cornstarch}, \blue{sugar}} \\ \cline{1-2}
$s_t$ & combine the sugar and cornstarch in a saucepan. &  & \\ 
\bottomrule
\end{tabular} }
\caption{Examples of selected ingredients. We provide the sentence $s_t$ and sentence $s_{t-1}$ for context. Ingredients are displayed in green if they are present in the ground truth set of ingredients and red otherwise. Best viewed in color.}
\label{tab:class_examples}
\vspace{-0.25cm}
\end{table*}

\subsection{Graph Generation}
In this section, we test the graph generation performance of our model. The goal is to validate the hypothesis that our proposed model is capable of generating meaningful graphs. For doing this, we compare against two different baselines and analyze the graphs obtained against the ground-truth graph-flow from  \cite{yamakata_english_2020}. 

\paragraph{Baselines} %The first baseline model we compare against is a rule based model making structural assumptions about the edge connectivity of each instruction in a given recipe. For this, we connect each node of type ingredient and location to a node to type action. Then, we assume the existence of an edge connecting the action of the instruction at time $t$ to the action at time $t+1$. This assumption is in line with the sequential nature of a cooking procedural text. 
The lack of large datasets with recipes and graph pairs is the main reason that motivates this study. In order to assess the validity of our method, we evaluate two aspects of our method. The first one is the quality of the generated text. This allows us to indirectly estimate whether the graph encoder is capable of extracting meaningful relations. We report the scores of this indirect evaluation using benchmark metrics for text generation. 
Additionally, we want to evaluate the quality of the graphs. For this, we use the dataset by  \citet{yamakata_english_2020}. We apply our model to the textual recipes and evaluate the text generation performance and the output graph in terms of graph edit distance (GED) \cite{sanfeliu_distance_1983}.  
For a novel line of work, the \textit{RandomGraph}, a random graph structure, is an intuitive baseline that gives us insights into our approach.
We include as a comparison the performance of our model without the concatenation of the sentence representations but only the node representation learned by the graph encoder. 

\paragraph{Metrics} For the graph $\rightarrow$ text task, we use BLEU \cite{papineni_bleu_2002} and ROUGE$_L$ \cite{lin_rouge_2004} to evaluate the closeness of our generated text to the input recipe. These metrics measure the accuracy of the n-grams between the predicted text and the ground truth. For evaluating the performance of the text $\rightarrow$ graph task, we measure the GED between our predicted graphs and the ground truth. This score is to find the best set of transformations that can transform graph $g_1$ into graph $g_2$ by means of edit operations on graph $g_1$. The allowed operations are inserting, deleting and/or substituting vertices and their corresponding edges:
\begin{equation}
 \mathsf{GED}(g_1, g_2) = \min_{e_1,...,e_k \in \gamma(g_1, g_2)}\sum_{i=1}^{k}c(e_i) 
 \vspace{-0.15cm}
\end{equation}
%$\mathsf{GED}(g_1, g_2) = \min_{e_1,...,e_k \in \gamma(g_1, g_2)}\sum_{i=1}^{k}c(e_i)$
where $c$ is the cost function for an editing operation $e_i$ and $\gamma(g_1, g_2)$ identifies the set of edit paths transforming $g_1$ into $g_2$.

\paragraph{Results} As shown in Table \ref{tab:gen_results}, our model achieves a good performance for the task of unsupervised graph generation. The recipes generated (an example is given in Table~\ref{tab:example_generated_text}) are coherent and follow the flow of the input recipes. Moreover, the analysis of some examples shows that in most cases the original ingredients are recognized by the model and placed in the correct context. Considering the graph generation, Table~\ref{tab:graph_edit_distance} shows the results of the GED score. We compare against the ground-truth flow graphs from the "EFC" corpus. It is important to clarify that these graphs are structurally different from ours. They are generated from a pre-processing task involving tagging each word of the recipe and predicting the edges between these nodes. Before the evaluation, we aggregate the duplicate nodes and discard input recipes with less than two steps. We then compute the $\mathsf{GED}$ by sampling 10 graphs and averaging the scores. This compromise is due to the high computational cost of the graph-matching problem. We compare the graphs obtained by our model to the processed ground truth and additionally compare the $\mathsf{GED}$ score to \textit{RandomGraph}. We can notice how the complete model including the sentence representation has a lower $\mathsf{GED}$ from the ground-truth set.

\section{Related Work}

\paragraph{Graph Generation} Graph generation is the task of modeling and
generating real-world graphs.
%, and it has applications in several domains, such as understanding interaction dynamics in social networks\cite{grover_graphite_2019,wang_graphgan_2018,tran_deepnc_2020}, anomaly detection \cite{ranu_graphsig_2009}, protein structure modeling \cite{guo_generating_2021,banerjee_proceedings_2022}, source code generation and translation \cite{du_interpretable_2022,brockschmidt_generative_2018,robins_recent_2007}, and semantic parsing \cite{zhang_amr_2019,dai_syntax-directed_2018}.
Early works generated graphs with structural assumptions defined a priori \cite{albert_statistical_2002, leskovec_kronecker_2010} although in many domains, the network properties and generation principles are largely unknown. Motivated by the need to improve fidelity, researchers studied methods to learn generative models from observed graphs. Recent works in graph generation exploit deep generative models such as variational autoencoders \cite{kingma_auto-encoding_2013} and generative adversarial models\cite{goodfellow_generative_2014}. There are different approaches for generating graphs and they differ based on the elements of the graph they generate at each step (or in one shot): (i) node \cite{su_graph_2019,assouel_defactor_2018,liu_constrained_2018,kearnes_decoding_2019}, edge \cite{goyal_graphgen_2020,bacciu_edge-based_2020,bacciu_graph_2019} and motif sequence \cite{liao_efficient_2019,podda_deep_2020}, respectively generate node, edge or sequence of smaller graphs at each time step; (ii) adjacency matrix based \cite{ma_constrained_2018,polsterl_likelihood-free_2019,de_cao_molgan_2018}, that directly learns a mapping to the latent embedding to
the output graph in terms of an adjacency matrix, generally with the addition of node/edge attribute matrices/tensors; and (iii) methods generate the edge
probability based on pairwise relationships between the
given nodes embedding \cite{grover_graphite_2019,kipf_variational_2016,shi_graphaf_2020} which is close to our method. The intuition behind these models is that nodes that are close in the embedding space should have a high probability of
being connected. However, we do not follow the standard method of enforcing semantic similarity between nodes as a mean for generating the adjacency matrix as the relationship we want to emerge from the graph is of the type \textit{action} $\rightarrow$ $\textit{ingredient}$ and \textit{location} $\rightarrow$ $\textit{action}$.

\paragraph{Text to graph} The literature on text-to-graph generation is predominantly based of knowledge graphs generation methods \cite{gardent_webnlg_2017,flanigan_generation_2016,konstas_inducing_2013,li_few-shot_2021} although other works generate flows graphs similar to ours. One specificity of our approach is that due to the generation methods, we do not allow for nodes to repeat themselves which is not the case for flow graphs. Additionally, our work relates to procedural knowledge extraction task \cite{qian_approach_2020,hanga_graph-based_2020,honkisz_concept_2018} which is fundamental for transforming natural texts into structured data like graphs. \cite{guo_cyclegt_2020} proposed an unsupervised training method that can iteratively back-translate between the text and graph data in a cyclic manner, such that the problem becomes similar to finding an alignment between unpaired graphs and related texts. \cite{guo_cyclegt_2020}. The fundamental difference is that we do not have access to graph-structure data.

\paragraph{Food understanding} Large datasets focusing on the cooking domain have gathered a new interest leading to advancements in food understanding. Most notable examples of these datasets are Food-101\cite{bossard_food-101mining_2014} and Recipe1M \cite{salvador_learning_2017}. Unlike our method, these benchmark datasets have been mostly exploited as reference benchmarks for the computer vision literature. The text-based studies in natural language processing have focused on tasks such as recipe generation in the context of procedural text understanding in the form of flow-graphs \cite{hammond_chef_1986,mori_flowgraph2text_2014,mori_flow_2014} and more generally recipe parsing \cite{chang_recipescape_2018,jermsurawong_predicting_2015,kiddon_mise_2015}. However, the datasets used in these studies are small (less than 300 recipes) and the supervision is limited. The edge connectivity is obtained from several structural assumptions and the nodes are uniquely identified within the graphs. 
%\red{We are targeting the same issues tackled by the previous works but at a larger scale}.
Our work proposed an unsupervised approach that provides a method to cope with scalability.

\paragraph{Conditional text generation} Conditional text generation with auto-regressive models has been widely studied in the literature. There are different types of conditioning. The neural machine translation task is an  example of text-based conditioning \cite{gehring_convolutional_2017,fan_hierarchical_2018,sutskever_sequence_2014,vaswani_attention_2017} is 
%, where the objective is to predict a translation in a different target language given a text in a source language.
The image captioning task \cite{vinyals_show_2015,xu_show_2015,lu_knowing_2017} uses conditioning based on data in other modalities, namely the input to the decoder is an image representation. Different architectures have been studied for this task, such as $\mathsf{RNN}$ \cite{fan_hierarchical_2018},  $\mathsf{CNNs}$ \cite{gehring_convolutional_2017} or attention-based methods \cite{vaswani_attention_2017}. In this paper, we exploit the conditional text generation task using the learned graph representation as a prompt to the text decoder.

\section{Conclusion}
We proposed a model that transforms a procedural text into a graph condensing the relevant information necessary to perform the procedures. This is performed in an unsupervised way as no supervision at the graph-level is provided. More specifically, only the raw recipe text is used to construct a graph. Experimental results show the validity of our approach. The objective relates to how faithful the recovered text is to the original recipe text. 
These representations are meant to able to provide abstraction from the expression in surface strings.
The graphs produced can be integrated into automated reasoning frameworks. In future work, we plan to generate richer graphs containing more information in terms of qualifiers and quantifiers, that can be used to teach automated agents how to execute and reason about procedures.

\section{Limitations}
The scope of this work is to study an approach to produce a formal representation in the form of graphs when no supervision data is available. The lack of large publicly available datasets and work studies focusing on this specific issue limits the depth of analysis that could be performed. Moreover, in this work, we did some specific choices that may make the graphs produced less directly human interpretable. This is highlighted by the choice of having persistent entities in the recipe represented by a single node that can have multiple connections whereas the flow graphs in the cooking domain stress less this aspect by allowing duplicates in the graph. This reflects the alternative intended use of the representations. 

There are additional concerns that need to be considered when generating graphs from cooking recipes. (i) bias: the model could be biased towards certain cuisines. Our model autonomously extracts the cooking ingredients to be included in the graphs. However, there is a possibility that some ingredients might not be recognized as such due to not being part of the dataset of origin used to train the entity selection module; (ii) inaccuracies or misinformation:  the graph generator may generate graphs that contain inaccurate or misleading information. This is due to the self-supervised nature of the framework.

Finally, future work involves an analysis of the semantic relations that are represented within the learned graphs to investigate whether programs can be predicted from this representation that could lead to a finished product.

\section{Ethics Statement}

\paragraph{Data sources} Now You're Cooking dataset \cite{bosselut_simulating_2018} and corpora used
in this work is publicly available. The English Flow Corpus dataset \cite{yamakata_english_2020} is made available upon requesting it from the original authors. To quantify the performance of our model, we use $\mathsf{BLEU}$ and $\mathsf{ROUGE_L}$ scoring implementations using the $\mathsf{pycocoevalcap}$ tool from \url{https://github.com/salaniz/pycocoevalcap}.
\paragraph{Models} The output graphs are closer to abstract representations and are not intended to be used by users to interact directly. The generative models are based on pre-trained language models, which may generate misleading information if not prompted correctly.

%\section*{Acknowledgements}

\bibliography{references}
\bibliographystyle{acl_natbib}

%\appendix

%\section{Example Appendix}
%\label{sec:appendix}

\end{document}